\documentclass[journal,twoside]{IEEEtran}
\usepackage{ifpdf}
\ifCLASSINFOpdf
  \usepackage[pdftex]{graphicx}
  \usepackage{graphbox}
  \graphicspath{{figures/}}
  \DeclareGraphicsExtensions{.pdf,.jpeg,.png}
\else
  \usepackage[dvips]{graphicx}
  \graphicspath{{figures/}}
  \DeclareGraphicsExtensions{.eps}
\fi
\usepackage{amsmath}
\usepackage[bookmarks=false]{hyperref}
\usepackage{booktabs}
\usepackage{siunitx}
\usepackage[table,dvipsnames]{xcolor}
\usepackage{cite}
\usepackage{nicefrac}

\newcommand{\ie}{i.\,e.\ }
\newcommand{\eg}{e.\,g.\ }

\begin{document}
\title{HED-UNet: Combined Segmentation and Edge Detection for Monitoring the Antarctic Coastline}

\author{Konrad~Heidler,
        Lichao~Mou,
        Celia~Baumhoer,
        Andreas~Dietz,
        and~Xiao~Xiang~Zhu,~\IEEEmembership{Fellow,~IEEE}%
\thanks{%
  This work is supported by the Helmholtz Association through the Helmholtz Information and Data Science Incubator project ``Artificial Intelligence for Cold Regions'', Acronym \emph{AI-Core}, by Helmholtz Association’s Initiative and Networking Fund through Helmholtz AI [grant number: ZT-I-PF-5-01] -- Local Unit ``Munich Unit @Aeronautics, Space and Transport (MASTr)'', and by the German Federal Ministry of Education and Research (BMBF)
  in the framework of the international future AI lab
  ``AI4EO -- Artificial Intelligence for Earth Observation: Reasoning, Uncertainties, Ethics and Beyond'' (Grant number: 01DD20001).
}%
\thanks{%
  K. Heidler, L. Mou and X. Zhu are with the Remote Sensing Technology Institute (IMF),
  German Aerospace Center (DLR), 82234 Wessling, Germany,
  and also with the Data Science in Earth Observation
  (SiPEO, formerly Signal Processing in Earth Observation),
  Technical University of Munich (TUM),
  80333 Munich, Germany.
  E-mails: konrad.heidler@dlr.de; lichao.mou@dlr.de; xiaoxiang.zhu@dlr.de}%
\thanks{%
  C. Baumhoer and A. Dietz are with the German Remote Sensing Data Center (DFD),
  German Aerospace Center (DLR), 82234 Wessling, Germany.
  E-mails: celia.baumhoer@dlr.de; andreas.dietz@dlr.de%
}%
}
\markboth{SUBMITTED TO IEEE TRANSACTIONS ON GEOSCIENCE AND REMOTE SENSING}%
{Heidler \MakeLowercase{\textit{et al.}}: Monitoring the Antarctic Coastline by Combining Semantic Segmentation and Edge Detection}

\maketitle
\begin{abstract}
  \textcolor{blue}{This work has been accepted by IEEE TGRS for publication.}
  Deep learning-based coastline detection algorithms
  have begun to outshine traditional statistical methods in recent years.
  However, they are usually trained only as single-purpose
  models to either segment land and water or delineate the coastline.
  In contrast to this, a human annotator will usually keep a mental map of both
  segmentation and delineation when performing manual coastline detection.
  To take into account this task duality,
  we therefore devise a new model to unite these two approaches in a deep learning model.
  By taking inspiration from the main building blocks of
  a semantic segmentation framework (UNet) and
  an edge detection framework (HED), both tasks are combined in a natural way.
  Training is made efficient by employing deep supervision on
  side predictions at multiple resolutions.
  Finally, a hierarchical attention mechanism is introduced to
  adaptively merge these multiscale predictions into the final model output.
  The advantages of this approach over other traditional and deep learning-based methods
  for coastline detection are demonstrated on a dataset of Sentinel-1 imagery
  covering parts of the Antarctic coast,
  where coastline detection is notoriously difficult.
  An implementation of our method is available at \url{https://github.com/khdlr/HED-UNet}.
\end{abstract}

\IEEEpeerreviewmaketitle
\section{Introduction}
\IEEEPARstart{C}{ontrary} to many other landmasses,
Antarctica's coastline is fringed by dynamic glacier and ice shelf fronts
continuously changing the coastline location by iceberg calving,
which is influenced by both seasonal variations as well as global climate change.
Tracking the advance and retreat of glacier and ice shelf fronts
is an important factor for a better understanding of glaciological processes.
Furthermore, it is essential to monitor calving front retreat
as it enhances the sea level contribution of the Antarctic ice sheet due to decreased buttressing effects.

Overall, the length of the Antarctic coastline amounts to around \SI{40000}{\kilo\metre}
\cite{liu_complete_2004}, which renders manual delineation infeasible.
Especially when observing the developments over multiple time steps
for continuous tracking, an automated coastline extraction technique is needed.
The recent advances in algorithms and sensing platforms
open up new possibilities for the analysis of satellite imagery over large regions,
which can be observed in fields as diverse as
land cover mapping~\cite{geng_deep_2017,robinson_large_2019,tong_land-cover_2020},
bathymetry~\cite{eugenio_high-resolution_2015,liang_derivation_2017,erena_bathymetry_2020},
urban applications~\cite{long_accurate_2017,audebert_beyond_2018,mou_vehicle_2018,mou_relation_2020,li_building_2020},
change detection~\cite{wen_novel_2016,ajadi_change_2016,lv_novel_2019,du_unsupervised_2019,zhang_detecting_2019},
and cryosphere research~\cite{engram_analyzing_2018,sasgen_high-resolution_2019,dammann_mapping_2019,nitze_remote_2018,anderson_linking_2019}.

This kind of fine-grained analysis is possible
because of the availability of satellite imagery with revisit times in the order of days.
Regarding data sources,
both optical and synthetic aperture radar (SAR) sensors
produce imagery suitable for the delineation of the Antarctic coastline~\cite{baumhoer_remote_2018}.
The use of optical imagery in the Antarctic comes with some major drawbacks.
Apart from the usual problems with cloud cover,
vision is further impeded by polar night and sensor saturation due to the high albedo of ice.
To create continuous and gapless observations,
data from the {Sentinel-1} mission was chosen as the main imagery source.
SAR data has often been found to be helpful with the analysis of the cryosphere~\cite{strozzi_glacier_2002,vasile_high-resolution_2008,erten_glacier_2013,akbari_monitoring_2014,lang_focused_2015,krieger_automatic_2017,akbari_iceberg_2018,baumhoer_automated_2019,zhang_automatically_2019,mohajerani_detection_2019}.
In our case, it allows for near-realtime analysis at a high temporal resolution.

Using SAR data for the task of coastline extraction also imposes some challenges.
The speckle present in SAR images makes it harder to pinpoint the exact boundary between land and sea.
Further, the backscatter characteristics of glacial ice vary throughout the year,
making it hard to distinguish between e.\,g. open sea and the higher ice sheet.
Therefore, a good model needs to pay additional attention to contextual clues
and cannot rely on local information only.

\begin{figure}
  \center
  \mbox{}\\[6pt]
  \setlength{\tabcolsep}{1pt}
  \renewcommand{\arraystretch}{1.4}
  \begin{tabular}{cccc}
    \includegraphics[width=0.27\linewidth,align=c]{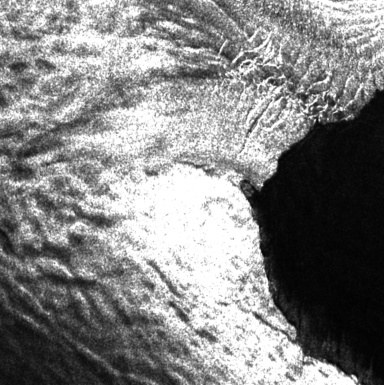}&
    \includegraphics[align=c]{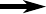}&
    \includegraphics[width=0.27\linewidth,align=c]{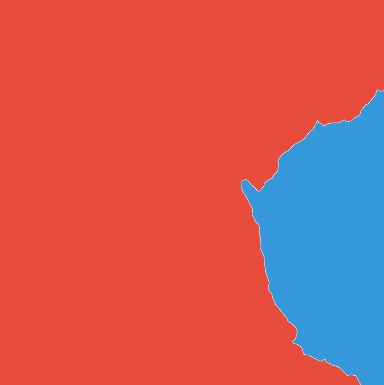}&
    \includegraphics[width=0.27\linewidth,align=c]{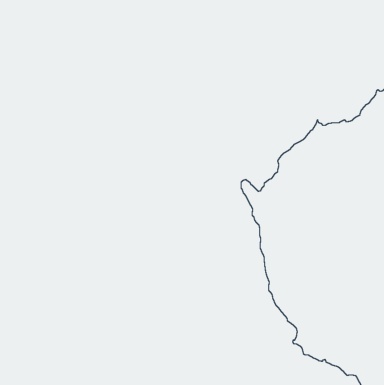}\\
    SAR imagery&&Segmentation&Edge Detection
  \end{tabular}
  \caption{In coastline detection, the vision tasks of segmentation and edge detection
  are inseparable.}%
\end{figure}

Existing studies for delineating coastlines in general,
as well as the Antarctic one,
often focus their predictions on either the
area of land and sea (\emph{sea-land segmentation}), or the coastline itself (\emph{coastline detection}).
But to the human eye, the two concepts of ``area'' and ``edge'' are closely intertwined,
making it hard to imagine one without the other.
When conducting manual coastline delineation,
a human annotator will therefore mentally segment the scene into sea and land
while searching for the edge between the two at the same time.

We hypothesize that taking into account this duality is essential
in closing the performance gap between human annotators and automated approaches.
In an attempt to more closely model this process,
we thus introduce a new solution for coastline detection
that draws upon the advantages of both segmentation and edge detection approaches.
Instead of focusing a predictor on just one of these tasks,
our network is trained to jointly perform both tasks at the same time.
Inspired by neural architectures for semantic segmentation and edge detection,
the model uses an encoder-decoder architecture with skip connections
in order to predict segmentation masks and edges at multiple resolutions.

Another observation we make about coastline detection conducted by humans
is the fact that not all areas of a given scene need the same amount of attention to detail.
While it is of paramount importance that the coastal regions are precisely mapped,
areas further away from the coastline do not receive much attention from a human annotator.
By introducing a merging scheme based on hierarchical attention,
our model can work in the same way.
The intermediate multiresolution predictions are merged using this mechanism
to obtain a final output that combines
fine-grained low level outputs with coarser high level outputs in an efficient way.

Overall, this work's contributions are threefold:
\begin{itemize}
  \item Coastline detection is recognized as a dual task.
    To solve this, a unified theory of segmentation and edge detection is presented.
    From this, an architecture that implements both semantic segmentation and
    edge detection is devised.
  \item Apart from the narrow coastal strip,
    there are large regions that require less detailed analysis.
    This is taken into account by allowing the model to output predictions at
    different resolution levels.
    Adding deep supervision for these side outputs improves the training efficiency
    and generalization performance of the model.
  \item In order to dynamically blend between coarse and high-resolution predictions,
    a hierarchical attention mechanism is used that takes into account the
    information available at all levels.
\end{itemize}

The remainder of the paper is organized as follows.
Section~\ref{sec:related} gives a brief overview of current methods for coastline detection
with a focus on polar regions,
as well as existing approaches for combining segmentation and edge detection.
Section~\ref{sec:method} presents our proposed HED-UNet architecture.
In Section~\ref{sec:experiments}, the used dataset is introduced.
Further, the conducted experiments are explained.
Finally, Section~\ref{sec:results} presents numerical results
comparing our model to other approaches and ablation studies that analyze the
proposed model's elements in detail. Finally, it also includes a discussion
of the observed model performance.

\section{Related Work}
\label{sec:related}
This section will explore the state of the art for coastline
detection with a focus on Antarctica.
Compared to the general case, the detection of coastlines in the Antarctic requires additional care,
as many methods are easily distracted by dynamic sea ice, like icebergs or ice m\'elange.
Locally, these confounding features can look almost identical to land ice,
and can therefore only be excluded by the additional use of spatial context information.

There are numerous existing approaches for detecting coastlines from satellite imagery.
For the biggest part, they can be divided into the aforementioned two classes,
differing in the output of interest.

\subsection{Sea-Land Segmentation}
In the field of computer vision, semantic segmentation is a central topic.
Each pixel is assigned a class which is to be predicted by the model.
This technique is frequently used in remote sensing for various tasks.
When the area of either sea or land is of importance,
semantic segmentation models are used to distinguish
between sea pixels and land pixels.

\subsubsection{Statistical Methods}
In quite a few studies, this has been done by means of statistical analysis.
For the Antarctic, the use of a bimodal Gaussian mixture model was proposed,
for which parameters are estimated in order to derive an adaptive thresholding scheme.
This approach can be applied to both SAR and optical imagery~\cite{liu_complete_2004}.
Similar dynamic thresholding schemes have been applied to different sensors~\cite{miles_simultaneous_2017}.
While easy to implement and fast to evaluate,
these methods completely discard the spatial relationships of the pixels,
which renders them unfit to deal with the aforementioned issues.

Another localized way of segmenting images that has been applied to sea-land segmentation
is given by the watershed algorithm~\cite{liu_ocean-driven_2015}.
It treats the pixel intensities as height values and then simulates the resulting surface being flooded with water.
Finally, unsupervised clustering methods are helpful in the analysis of complex coastlines~\cite{schmitt_potential_2019}.
These methods have the benefit of being unsupervised, \ie requiring no training
prior to the evaluation,
but the lack of supervision also means that the models cannot be taught to \eg ignore icebergs.

\subsubsection{Deep Learning Methods}
With the rise of deep learning in remote sensing \cite{zhu2017dl},
convolutional neural networks (CNNs) have been shown to provide superior performance for many tasks,
including the one of sea-land
segmentation~\cite{shamsolmoali_novel_2019,li_deepunet_2018,chu_sea-land_2019}.
Deep convolutional architectures
like SegNet~\cite{badrinarayanan_segnet_2017}
or UNet~\cite{ronneberger_u-net_2015}
leverage contextual information through their encoder-decoder architectures.
So as they have more context to base their decisions on,
they have the potential to produce more accurate results than pixelwise
or shallow texture-based classifiers.
This is of great interest to Antarctic coastline detection due to the aforementioned issues.
Current developments in computer vision
show a trend towards more complex models for semantic segmentation,
which incorporate global information~\cite{yuan_object-contextual_2020}
or shape information~\cite{takikawa_gated-scnn_2019}.

Generally, these models require large amounts of labeled data,
and take quite some time to train.
However, they can outperform the previously mentioned methods.

\subsection{Coastline Detection}
A closely related task is approached in coastline detection.
Instead of segmenting a scene into sea and land,
the coastline itself is of primary interest.

\subsubsection{Edge Tracing}
One class of edge detection methods
mark the boundaries in the image step by step.
After some filtering to highlight the edges,
which can be done \eg
using the Roberts operator~\cite{roberts_machine_1963}
or the Sobel operator~\cite{pratt_edge_2006},
pixels that are likely to lie on the edge are connected to form the
entire boundary.
Regarding coastline detection, this approach has been shown to work for SAR data,
when applying preprocessing steps to account for the nature of the imagery~\cite{lee_coastline_1990}.
They can also be connected using a shortest-path algorithm~\cite{krieger_automatic_2017},
or ridge tracing~\cite{wang_coastline_2019}.
Yet another approach comes from exploiting detection duality.
By the nature of the relation between sea, land and the coastline,
the coastline can be derived from a sea-land segmentation by tracing the transitions
between the sea and land class~\cite{modava_integration_2019}.

While relatively simple,
these methods often have some issues regarding robustness.
When the tracing procedure takes a wrong turn,
it is hard for the algorithm to return to the true boundary.

\subsubsection{Contour Methods}
Active contours,
sometimes also called Snakes~\cite{kass_snakes_1988},
are quite similar to the edge tracing approach.
Instead of the pixel-by-pixel approach,
this class of methods uses an initial curve
that is iteratively deformed to minimize an energy function.
By choosing the right energy function,
this framework can be used to delineate coastlines.
For SAR imagery,
active contours are able to find coastlines when given a
good initialization~\cite{klinger_antarctic_2011,liu_coastline_2017}.
These models are sensitive to the provided initialization,
meaning that they can converge to local minima that do not represent the desired edge.

\subsubsection{Level Set Methods}
Instead of working with an explicit parametrization of the curve,
  these methods work with an implicit representation given by a scalar field
  in which the zero set represents the boundary~\cite{osher_fronts_1988,chan_active_1999}.
Adaptations of this method for SAR coastline detection use
multiple level set iterations to go from coarse to fine delineations~\cite{liu_coastline_2016}
or sophisticated preprocessing steps~\cite{modava_level_2017} to
make the method work for this particular type of imagery.

\subsubsection{Deep Learning Methods}
Only recently have approaches based on deep learning begun to outperform
handcrafted edge detection algorithms.
Specialized architectures leverage the framework of CNNs to derive
features that predict the presence of edges%
~\cite{xie_holistically-nested_2015,liu_richer_2017,poma_dense_2020}.
Notably, the previously mentioned Roberts and Sobel operators
can be viewed as a shallow CNNs with just one layer and a convolutional filter size of 2 and 3 respectively.
Therefore, it is only natural that deeper CNNs with more layers are able to outperform these
hardcoded edge detection operators.

\subsection{Combining Semantic Segmentation and Edge Detection}
A common problem with semantic segmentation models is the blurriness near class boundaries.
This likely stems from the fact that the edges make up a minority of the pixels,
and are therefore not well enough represented by the standard pixelwise cross-entropy loss.
Thus, the idea of augmenting semantic segmentation approaches
with edge information is not a new one.

One way of making a segmentation model aware of edges in the image
is by adding an auxiliary loss term that encourages the prediction
of crisp edges.
This has been shown to work for sea-land segmentation~\cite{cheng_senet_2017}.

Surprisingly,
simply adding the edge detection task as an auxiliary output for a segmentation model
can improve the segmentation results quite a bit,
even without further changes to the model~\cite{jiang_semantic_2020}.
This approach can also improve sea-land segmentation results
in harbor areas~\cite{cheng_fusionnet_2017}.

To further improve blurry segmentations,
edge masks can be used as the basis for
a spatial propagation of class labels.
In~\cite{chen_semantic_2016}, a segmentation map is initialized using a segmentation network
and at the same time, edges are predicted.
These edge masks are then used as the basis for a recursive multidirectional label propagation. 

For aerial scene classification, the use of an edge detection subnetwork before doing the segmentation
has been shown to be beneficial.
The detected edge masks are then used as additional input features for the segmentation model.
This approach improves the shape accuracy of the resulting segmentation~\cite{marmanis_classification_2018}.

Contrary to these approaches,
we develop a unified theory of segmentation and edge detection.
We then identify the components that successful neural networks
use to solve either one of these tasks,
and finally devise a model that incorporates the tools necessary 
to solve both tasks at the same time.
The underlying assumption is that both segmentation and edge detection
are of equivalent importance for detecting coastlines in satellite imagery. 

\section{Proposed Method}\label{sec:method}
Implementing the sea-land segmentation task via a UNet segmentation model \cite{ronneberger_u-net_2015}
has become a popular approach for the automatic delineation of
coastlines~\cite{shamsolmoali_novel_2019,li_deepunet_2018,chu_sea-land_2019}.
And also on our dataset, this method yields good results on the majority of the evaluated
scenes~\cite{baumhoer_automated_2019}.
But oftentimes the predictions become inaccurate and blurry
in areas close to the coastline.
As the precise location of the coastline is the central object of our study,

On the other hand, edge detection models excel at delineating the edges in the given images.
However, an edge delineation has no concept of ``inside'' and ``outside'' by itself,
so this output alone is insufficient for labeling sea and land.
Further, edge detection models are easily fooled by inland
structures of similar appearance to the coastline, as well as icebergs near the coast.
This implies the need for extensive post-processing and manual corrections.

To put our aforementioned hypotheses into practice, 
we now introduce a hybrid model for simultaneous prediction of the sea-land segmentation
and edge detection of the coastline.
Following our observation that humans will usually take into account both the edge information
as well as the textural shape information,
we therefore propose a combined framework that draws upon 
the advantages of both these approaches.
It takes inspiration from both UNet~\cite{ronneberger_u-net_2015}
and HED~\cite{xie_holistically-nested_2015}, as well as related architectures
by combining key ideas in a very natural way.
Therefore, we call our model HED-UNet.

\subsection{Unifying Segmentation and Edge Detection}
\begin{figure}
  \centering
  \includegraphics[width=\linewidth]{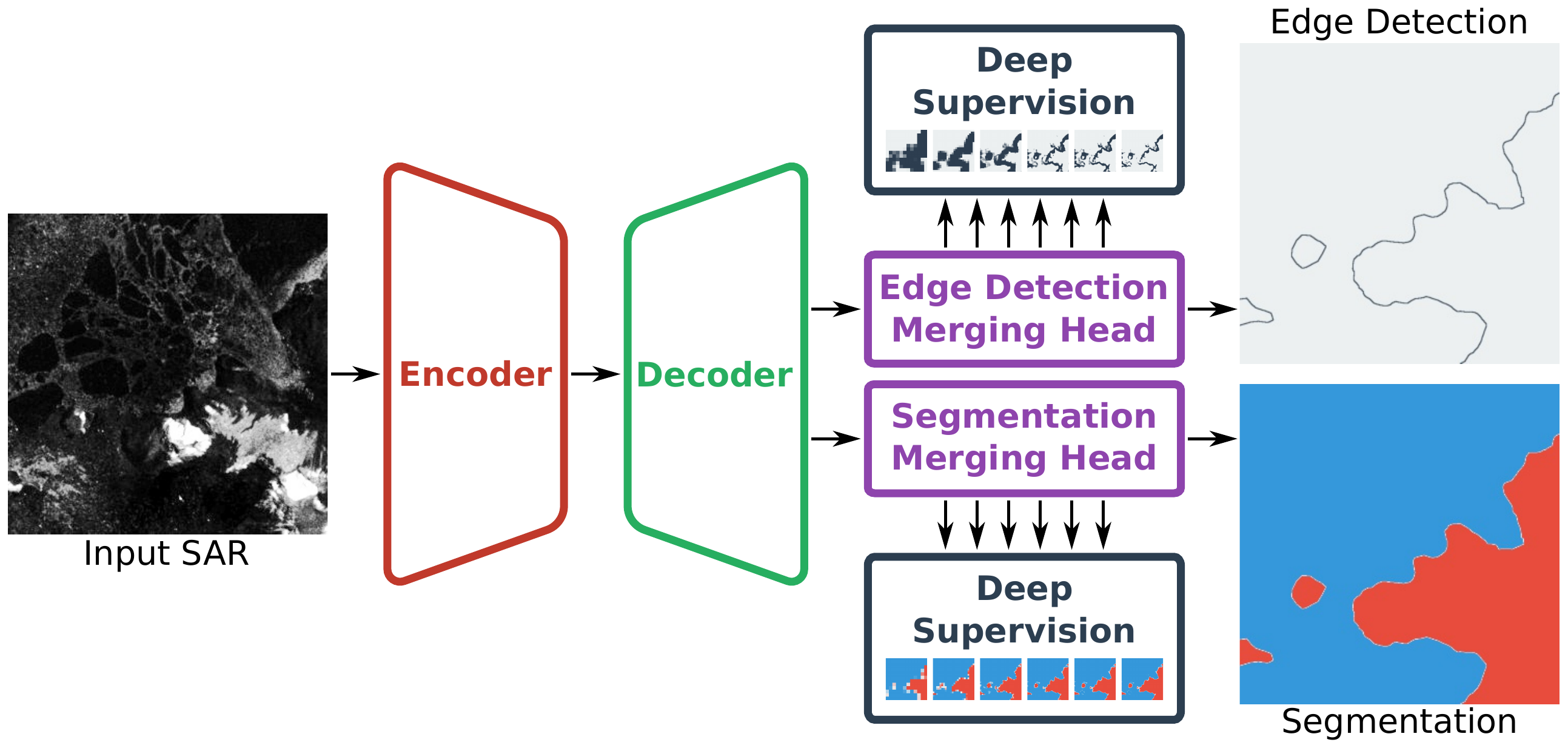}
  \caption{High-level structure of the proposed framework.
    First, the encoder and decoder calculate a pyramid of feature maps.
    Then, the task-specific merging heads combine this information
    using the hierarchical attention mechanism.}\label{fig:highlevel}
\end{figure}
\begin{figure*}
  \centering
  \includegraphics[align=t,width=\linewidth]{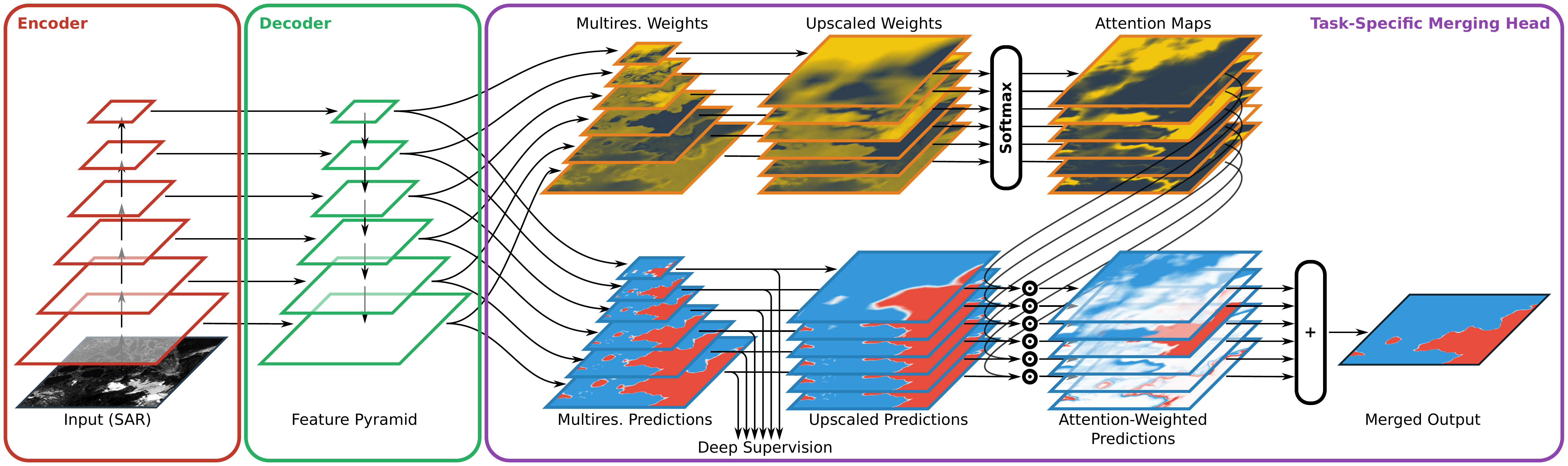}
  \caption{
    Architectural details of the proposed network.
    The full model contains two task-specific merging heads,
    for clarity, only the segmentation head is shown here.
    The edge detection head follows the same structure.}
  \label{fig:architecture}
\end{figure*}
Regarding the deep learning formulation of the tasks,
both segmentation and edge detection are in their nature \emph{dense prediction tasks},
\ie for each input pixel, an output label needs to be predicted.
In the case of segmentation, this is the class label, like ``sea'' or ``land''.
For edge detection, it is a classification into the two classes ``edge'' and ``no edge''.

This means that, in principle,
a segmentation model can be trained to perform edge detection and vice versa.
However, these models were designed for their respective tasks only,
meaning the performance will be degraded when applying them to a different task.
In order to construct a model that works well for both tasks,
we will therefore identify the components of successful architectures for both tasks,
and find a way to incorporate them into a single multitask model.

\subsubsection{Segmentation Building Blocks}
Some successful semantic segmentation architectures
employ the combination of an \emph{encoder} and a
\emph{decoder}~\cite{ronneberger_u-net_2015,badrinarayanan_segnet_2017}.
The encoder conducts a series of downsampling steps to
allow for the aggregation of contextual information at a lower resolution.
In turn, the decoder then distributes this information to the individual pixels
through a series of upsampling steps.

In a more recent branch of semantic segmentation approaches,
the network architecture is divided into a \emph{backbone} network that
calculates feature maps,
and one or multiple \emph{prediction heads}, which conduct the final classification
based on these feature maps~\cite{chen_rethinking_2017,takikawa_gated-scnn_2019,%
yuan_object-contextual_2020}.

The contextual aggregation capabilities of an encoder-decoder framework are needed for
this task, as some regions can only be classified correctly by the use of contextual clues.
At the same time, the backbone-head approach makes it easy
to build models that tackle multiple tasks.
These considerations lead to the idea of implementing
backbone network that follows the encoder-decoder structure.
This has been pioneered for the task of object detection
in the framework of \emph{feature pyramid networks}~\cite{lin_feature_2017}.
For our network, we will employ two task-specific prediction heads 
after calculating a feature pyramid through an encoder-decoder approach.

\subsubsection{Edge Detection Building Blocks}
On the other hand,
edge detection frameworks are optimized to provide sharp edge delineations
while at the same time keeping down the amount of false positives.
This means that they need to combine the crisp edges predicted at a high resolution
with more robust, lower resolution features to reject false positives from the former.
Edge detection methods therefore often try to strike a balance between
predictions or feature maps at different resolutions,
which can be done with an architecture that employs an encoder followed
by a merging block~\cite{xie_holistically-nested_2015,liu_richer_2017,poma_dense_2020}.
The encoder part is similar to the encoders used in semantic segmentation models,
it aggregates contextual information by downsampling.
The merging part however is a new block that combines the information from different
resolution levels after they have been upsampled to the full resolution.

Looking back at the proposed feature pyramid backbone,
such a merging part fulfills the function of a prediction head.
This observation leads to the high level network architecture,
as shown in Fig.~\ref{fig:highlevel}.
It is structured in such a way that it contains the components
for both a segmentation and an edge detection network.
After this general structure of the network has been fixed,
the detailed layout for each one of these blocks will be outlined in Section~\ref{sec:details}.

\subsubsection{Loss Function}
In edge detection, the classes ``edge'' and ``no edge''
are highly imbalanced.
Therefore, we use an adaptively balancing modification of
the binary cross-entropy loss, as proposed in~\cite{xie_holistically-nested_2015}.
For a single image with
a ground truth partition into positive pixels $Y_+$
and negative pixels $Y_-$ and a prediction $\hat p$,
it is given as

\begin{equation}
  \mathcal{L}(\hat p)
  =
  -\frac{|Y_-|\sum_{j \in Y_+} {\log \hat p_j}}{|Y_+ \cup Y_-|}
  -\frac{|Y_+|\sum_{j \in Y_-} {\log (1-\hat p_j)}}{|Y_+ \cup Y_-|}\,.
\end{equation}

This loss function gives equal weight to the positive and negative classes,
no matter the ratio between the two class sizes.
Thanks to this property, it is fit not only for edge detection,
but for semantic segmentation as well.
Therefore, it is used as the loss function for both tasks.

\subsection{Architecture Details}\label{sec:details}
Regarding the model details, we start with the encoder-decoder backbone.
Conjecturing that the model needs a large spatial context window to base its decisions on,
we use a feature pyramid with 6 resolution levels,
corresponding to 5 down- and upsampling steps.
In this pyramid, the finest feature map is at the full image resolution,
and the coarsest one is at $\nicefrac{1}{32}$ the resolution.
The number 6 was chosen to cover a large enough receptive field needed for the task.
Deepening the network even further would lead to
receptive fields that exceed the image tiles' extents,
and did not bring further improvements in our experiments.
In the decoder part, the data flows are merged by element-wise addition.

Inspired by the hierarchical nature of the HED architecture~\cite{xie_holistically-nested_2015},
we adopt the scheme of predicting coarse representations of the output from within deeper layers.
A side output for both segmentation and edge detection is added
for each feature map, for a total of 6 outputs. 
These multiscale outputs are used in two different ways.

\subsubsection{Deep Supervision}
When building a deep feature pyramid like here,
there might not be much motivation for the model to
encode meaningful and informative features to the deep, lowest resolution feature maps.
In order to explicitly provide this motivation,
we train the model to be able to predict the ground truth from
each single feature map in the pyramid.

This so-called \emph{deep supervision}~\cite{lee_deeply-supervised_2015}
is known to improve the learning effectiveness of a neural network,
as well as its generalization capabilities.
This is achieved by training intermediate network outputs on the ground truth data
to provide additional and more direct training feedback to the earlier layers.
In our case,
an accordingly downsampled version of the ground truth segmentation is created
for each one of the multiresolution predictions,
and the corresponding edges are calculated.
Then, these multiscale ground truths are compared with the predictions to provide additional loss terms.
The resulting deep supervision encourages the network to better capture larger structures
and make use of the available receptive field by encoding meaningful features in the deep layers.

\subsubsection{Multiscale Fusion}
In the next step, these side outputs become part of the merging heads
that combine the intermediate outputs into one full-resolution prediction.
This is a central point in the original HED architecture~\cite{xie_holistically-nested_2015},
so we also implement it in the combined HED-UNet model.
In this way, the model has a way of combining
fine-grained delineations near the edges with the
more robust high-level predictions further away from the edge.
The way of merging used in HED is to combine the
intermediate predictions using learned weights.
But to further improve the merging performance,
we propose the following attention-based merging mechanism.

\subsection{Hierarchical Attention Merging Heads}\label{sec:attn_merging}
The final element of the network architecture are the merging heads.
In the edge detection frameworks introduced
earlier~\cite{xie_holistically-nested_2015,liu_richer_2017,poma_dense_2020},
this is done by feature-wise concatenation, followed by a 1$\times$1 convolution
to merge the information from different levels.
But in different areas, different fusion behavior might be needed.
In coastal areas, the model might want to use predictions of the highest possible resolution
in order to accurately delineate the coastline.
However, farther away from the coast the lower resolution levels can provide a
more general assessment of the scene, and thus lead to better classifications in these areas.

To allow for this adaptive fusion of the multiscale predictions
that takes into account the confidence at the different granularities,
we therefore introduce a new fusion procedure based on \emph{attention}.
This technique was initially explored in natural language processing 
as sequential attention among words and tokens~\cite{vaswani_attention_2017},
and later also applied in computer vision as spatial attention within an image~\cite{wang_non-local_2018}. 

Inspired by these works, we apply attention for merging multiscale predictions.
Here, this mechanism allows the network to focus on the features that it deems most useful
for each pixel of the current scene, instead of having fixed weights for feature fusion.
So instead of sequential or spatial attention,
our attention block allows the model to \emph{attend to different resolution levels}.
It works like this:

For each prediction level, a weight map is created.
The weight maps are then upsampled to match the output resolution,
and turned into a categorical probability map by applying the softmax function over the
concatenated resolution levels.
To obtain the final prediction, the dot product between the predictions
and the attention mask is calculated.
This process is visualized in Fig.~\ref{fig:architecture}.

For a pyramid of feature maps $F_k$,
the final prediction $\hat p$ is thus calculated as:
\begin{equation}
  \hat p = \sum_{k}
  \mathrm{u}\big(f_k(F_k)\big) \cdot \mathrm{softmax}_k\Big(
    \mathrm{u}\big(g_k(F_k)\big)
  \Big)\,,
\end{equation}
where $\operatorname{u}(\cdot)$ denotes bilinear upsampling to the full output resolution.
The functions $f_k$, $g_k$ denote the multilevel prediction layers and the attention layers respectively,
both are implemented as simple $1 \times 1$ convolutional layers.

This approach can be interpreted probabilistically as follows.
The intermediate predictions $f_k(F_k)$ can be considered to be maps of
Bernoulli probabilities for the output classification at different resolutions.
Through the prediction process, these probabilities are conditioned on the input imagery.
The original merging procedure with fixed weights corresponds to a mixture
model of these Bernoulli maps where the mixture coefficients $w_k$ are learned and fixed.
For an input scene $X$, the predicted probabilities $Y$ are thus approximated as

\begin{equation}
  P(Y_{ij} \mid X) \approx
  \sum_{k} w_k\,P(Y_{ij} \mid X, \mathrm{resolution} = k)\,.
\end{equation}

Contrary to that, the attention merging corresponds to a mixture model where the mixture coefficients
$w_{kij}$ are learned to dynamically depend on the input as well,
resulting in the slightly different approximation
\begin{equation}
  P(Y_{ij} \mid X) \approx
  \sum_{k} w_{kij}(X)\,P(Y_{ij} \mid X, \mathrm{resolution} = k)\,.
\end{equation}

Notation-wise, this might seem like a small change.
However, it leads to more flexibility in the resulting probabilistic model,
which implies the potential for better classifications.

From the probabilistic perspective,
the model training corresponds to a simultaneous
maximization of both the side outputs' likelihood as well as the likelihood of the full mixture under
the observed data.

\section{Dataset and Experimental Setup}
\label{sec:experiments}
\begin{figure}
  \center
  \includegraphics[width=\linewidth]{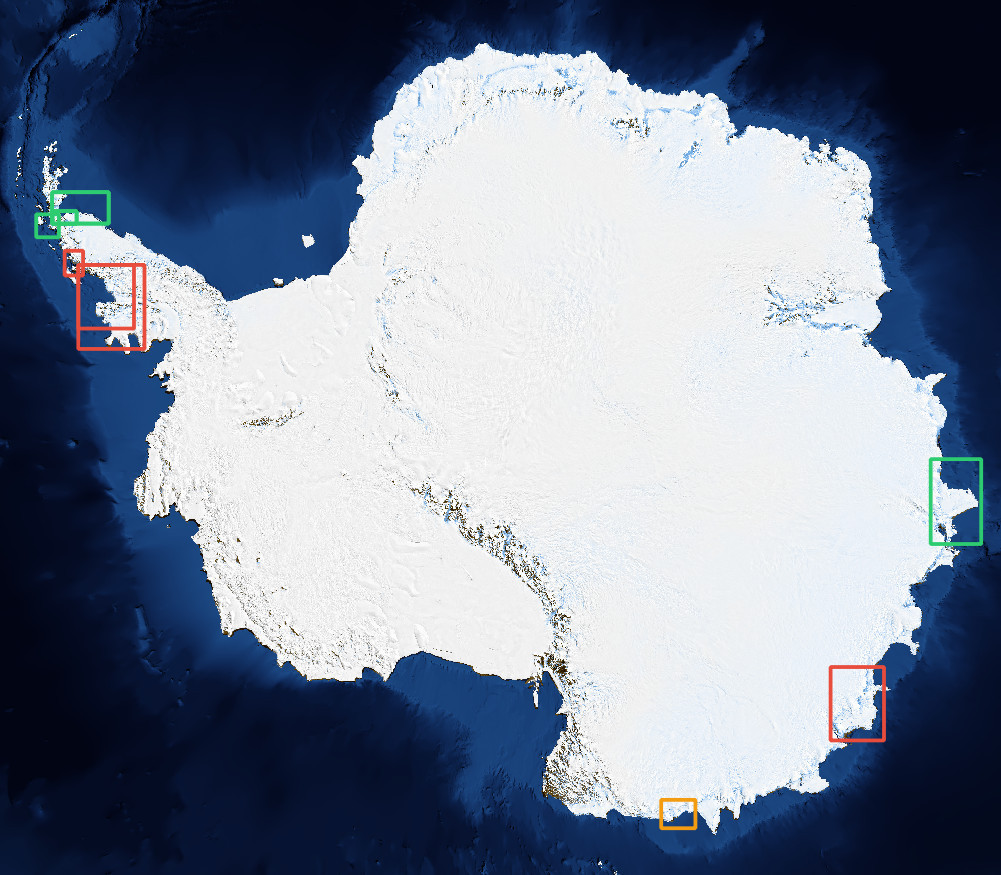}
  \caption{Spatial distribution of the scenes in the dataset.
  Scenes marked in green were used for model training,
  scenes marked in red were used for validation purposes.
  The red area in the top left is the ``Antarctic Peninsula'' validation site,
  while the bottom right red area is the ``Wilkes Land'' validation site.
  For most locations, data from 2 or 3 different sensing dates was used to
  allow for an assessment of each model's temporal stability.
  Marked in yellow is the footprint of the visualization tile in Fig.~\ref{fig:largevis}.
  }\label{fig:sites}
\end{figure}

\begin{figure*}
  \newlength{\tilewidth}
  \setlength{\tilewidth}{0.11\linewidth}
  \setlength{\tabcolsep}{1pt}
  \renewcommand\arraystretch{0.7}
  \newcommand{\example}[1]{%
    \includegraphics[width=\tilewidth]{figures/autotiles/#1-sar.jpg}&%
    \includegraphics[width=\tilewidth]{figures/autotiles/#1-seg-LiuJezek.jpg}&
    \includegraphics[width=\tilewidth]{figures/autotiles/#1-seg-UNet-1.jpg}&%
    \includegraphics[width=\tilewidth]{figures/autotiles/#1-seg-RDUNet-1.jpg}&%
    \includegraphics[width=\tilewidth]{figures/autotiles/#1-seg-GSCNN-1.jpg}&%
    \includegraphics[width=\tilewidth]{figures/autotiles/#1-seg-HRNetOCR-1.jpg}&%
    \includegraphics[width=\tilewidth]{figures/autotiles/#1-seg-HED-UNet-S-1.jpg}&%
    \includegraphics[width=\tilewidth]{figures/autotiles/#1-seg-gt.jpg}&%
    \\
  }%
  \center
  {\large\textbf{Segmentation}}\\[0.5em]
  \begin{tabular}{ccccccccccccc}
    \scriptsize{}SAR&%
    \scriptsize{}Gauss. Mixture~\cite{liu_complete_2004}&
    \scriptsize{}UNet~\cite{ronneberger_u-net_2015}&%
    \scriptsize{}RDUNet~\cite{shamsolmoali_novel_2019}&%
    \scriptsize{}Gated-SCNN~\cite{takikawa_gated-scnn_2019}&
    \scriptsize{}HRNet+OCR~\cite{yuan_object-contextual_2020}&%
    \scriptsize{}HED-UNet&%
    \scriptsize{}Ground Truth%
    \\
    \example{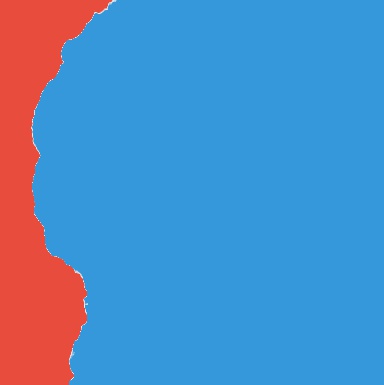}
    \example{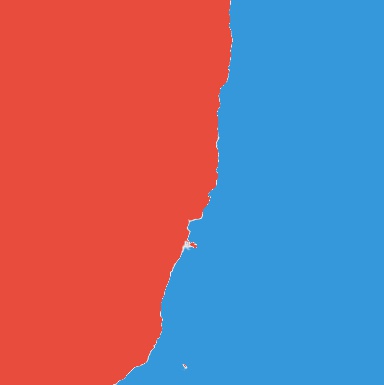}
    \example{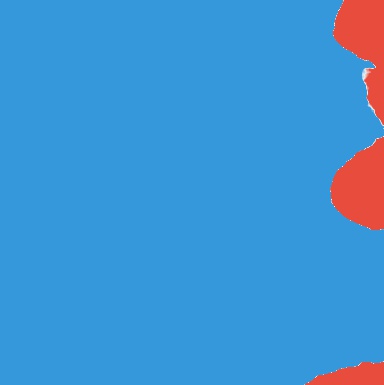}
    \example{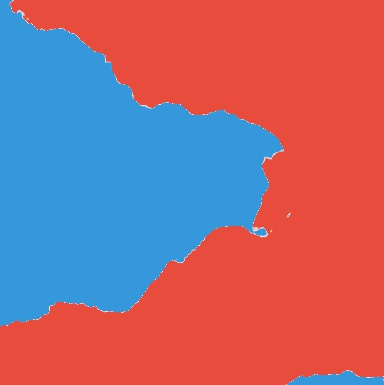}
    \example{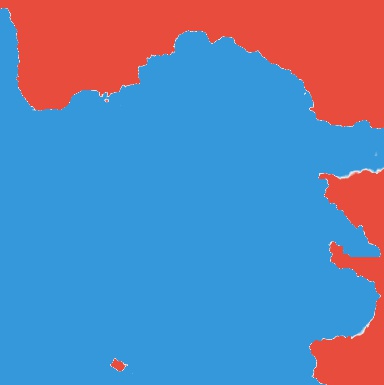}
  \end{tabular}\\
  \renewcommand{\example}[1]{%
    \includegraphics[width=\tilewidth]{figures/autotiles/#1-sar.jpg}&%
    \includegraphics[width=\tilewidth]{figures/autotiles/#1-edge-ActiveContour.jpg}&%
    \includegraphics[width=\tilewidth]{figures/autotiles/#1-edge-Lee.jpg}&%
    \includegraphics[width=\tilewidth]{figures/autotiles/#1-edge-HED-1.jpg}&%
    \includegraphics[width=\tilewidth]{figures/autotiles/#1-edge-GSCNN-1.jpg}&%
    \includegraphics[width=\tilewidth]{figures/autotiles/#1-edge-HED-UNet-S-1.jpg}&%
    \includegraphics[width=\tilewidth]{figures/autotiles/#1-edge-gt.jpg}&%
    \\
  }%
  \center
  {\large\textbf{Edge Detection}}\\[0.5em]
  \begin{tabular}{ccccccccccccc}
    \scriptsize{}SAR&%
    \scriptsize{}Active Contour~\cite{chan_active_1999}&%
    \scriptsize{}Sobel~\cite{lee_coastline_1990}&%
    \scriptsize{}HED~\cite{xie_holistically-nested_2015}&
    \scriptsize{}Gated-SCNN~\cite{takikawa_gated-scnn_2019}&
    \scriptsize{}HED-UNet&%
    \scriptsize{}Ground Truth%
    \\
    \example{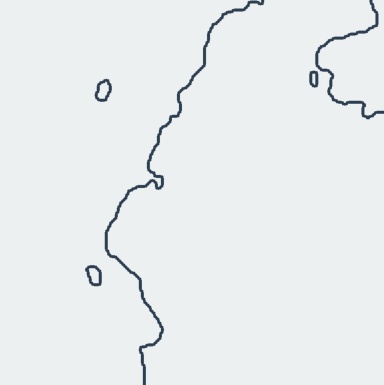}
    \example{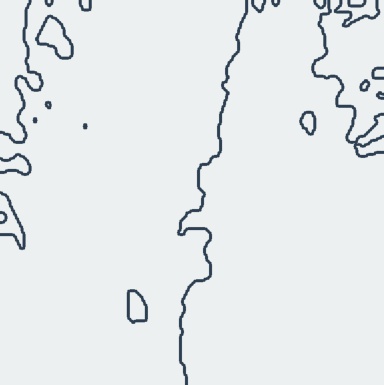}
    \example{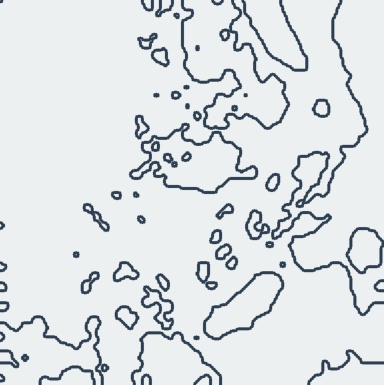}
    \example{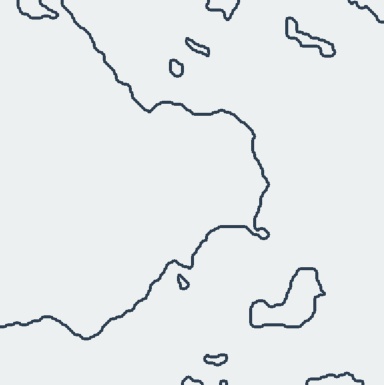}
    \example{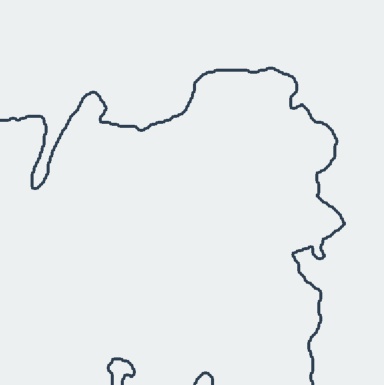}
  \end{tabular}\\
  \caption{Qualitative results comparing the evaluated models on unseen validation tiles.
    In order to provide an informative visualization,
    the visualized tiles were selected to represent the full spectrum
  of easy (top) to hard (bottom) scenes within the validation set.
    }\label{fig:qualitative}
\end{figure*}
\begin{table*}
  \center
  \caption{%
    Numerical Results for the Evaluated Models
  }\label{tab:results}%
  \setlength\tabcolsep{3pt}
  \vspace{-2em}
  \footnotesize
  \begin{tabular}{lllllllllll}
\toprule
Site & \multicolumn{5}{c}{Wilkes Land} & \multicolumn{5}{c}{Antarctic Peninsula} \\
                               Metric             &          Accuracy &              mIoU &                 Deviation &          F$_1$ ODS &         F$_1$ OIS &          Accuracy &              mIoU &                  Deviation &         F$_1$ ODS &         F$_1$ OIS \\
\midrule
Gaussian Mixture~\cite{liu_complete_2004}         &              77.4 &              63.0 &                       773 &                    &                   &              74.7 &              58.8 &                        765 &                   &                   \\
K-Medians Clustering~\cite{schmitt_potential_2019}&              55.9 &              28.0 &                       637 &                    &                   &              60.5 &              40.1 &                        560 &                   &                   \\
Sobel Edges~\cite{lee_coastline_1990}             &                   &                   &                       507 &               29.0 &              31.8 &                   &                   &                        644 &              21.1 &              20.8 \\
Active Contours~\cite{chan_active_1999}           &                   &                   &                       672 &               21.9 &              23.5 &                   &                   &                        698 &              14.6 &              15.1 \\
HED~\cite{xie_holistically-nested_2015}           &                   &                   &            341$\,\pm\,$22 &   38.4$\,\pm\,$1.7 &  41.0$\,\pm\,$1.0 &                   &                   &  398$\,\pm\,$\phantom{0}27 &  28.5$\,\pm\,$0.8 &  29.6$\,\pm\,$0.7 \\
UNet~\cite{ronneberger_u-net_2015}                &  89.2$\,\pm\,$3.0 &  80.6$\,\pm\,$4.7 &            271$\,\pm\,$14 &                    &                   &  79.3$\,\pm\,$2.8 &  65.0$\,\pm\,$4.4 &  483$\,\pm\,$\phantom{0}40 &                   &                   \\
DeepUNet~\cite{li_deepunet_2018}                  &  87.3$\,\pm\,$6.4 &  77.6$\,\pm\,$9.9 &            287$\,\pm\,$32 &                    &                   &  76.9$\,\pm\,$4.8 &  61.8$\,\pm\,$7.5 &            525$\,\pm\,$118 &                   &                   \\
RDUNet~\cite{shamsolmoali_novel_2019}             &  89.2$\,\pm\,$1.4 &  80.1$\,\pm\,$2.2 &            271$\,\pm\,$26 &                    &                   &  78.3$\,\pm\,$1.2 &  63.9$\,\pm\,$2.0 &  460$\,\pm\,$\phantom{0}73 &                   &                   \\
HRNet + OCR~\cite{yuan_object-contextual_2020} &  89.2$\,\pm\,$2.5 &  80.2$\,\pm\,$4.3 &            262$\,\pm\,$35 &                    &                   &  78.6$\,\pm\,$1.9 &  64.6$\,\pm\,$2.5 &  467$\,\pm\,$\phantom{0}61 &                   &                   \\
Gated-SCNN~\cite{takikawa_gated-scnn_2019}        &  87.1$\,\pm\,$0.2 &  76.8$\,\pm\,$0.1 &  297$\,\pm\,$\phantom{0}2 &   31.6$\,\pm\,$0.4 &  34.1$\,\pm\,$0.1 &  77.7$\,\pm\,$1.5 &  63.0$\,\pm\,$2.2 &  471$\,\pm\,$\phantom{0}33 &  23.0$\,\pm\,$1.7 &  25.4$\,\pm\,$1.7 \\
HED-UNet                                          &  92.0$\,\pm\,$0.8 &  84.9$\,\pm\,$1.4 &            222$\,\pm\,$23 &   39.7$\,\pm\,$1.2 &  41.6$\,\pm\,$0.9 &  80.5$\,\pm\,$1.6 &  67.2$\,\pm\,$2.2 &  345$\,\pm\,$\phantom{0}24 &  27.1$\,\pm\,$1.9 &  29.4$\,\pm\,$1.8 \\
\bottomrule
\end{tabular}

\end{table*}

In order to validate the effectiveness of the suggested improvements,
we trained and validated several competing methods as well as the proposed model
on a dataset of the Antarctic coast.

\subsection{Dataset}
Our dataset consists of 16 cropped Sentinel-1 GRD scenes of Antarctica's coastline
taken between June 2017 and December 2018 in the sensor's Extra Wide Swath acquisition mode.
The spatial distribution of these tiles can be seen in Fig.~\ref{fig:sites}.
The data has a resolution of 40\,m and dual polarization with HH and HV channels.
The cropped scenes have an average size of $7870 \times 6572$ pixels
($\SI{315}{\kilo\metre} \times \SI{263}{\kilo\metre}$),
and a combined area of around \SI{730000}{\kilo\metre^2}.
All imagery is processed 
in the Antarctic Polar Stereographic projection (EPSG:3031)
and converted to decibel.
On these scenes,
the coastline was manually annotated by experts in order to provide a
ground truth sea-land segmentation and coastline delineation.

The scenes within the dataset are clustered in 4 areas,
out of which 2 were selected as validation areas and
completely left them out of the training procedure.
This leads to a split of 11 training scenes and 5 validation scenes.
The scenes were all tiled into sections of $768 \times 768$ pixels
with 50\% overlap between adjacent tiles to form the training and
validation dataset, respectively.
In order to improve generalization performance,
we employed 8-fold data augmentation on the training set.
This augmentation technique processes a single tile into the
8 different versions that can be obtained by horizontal or vertical mirroring,
as well as rotating by multiples of 90$^\circ$.

\subsection{Evaluated Models}
As competitors to our model we evaluate the following models
to provide a baseline.
\subsubsection{Traditional Methods}\mbox{}\\
\textbf{Gaussian Mixture}
The sea-land segmentation method presented in~\cite{liu_complete_2004},
which applies dynamic thresholding based on a bimodal mixture of gaussians.\\
\textbf{K-Medians Clustering}
An unsupervised sea-land segmentation method presented in~\cite{schmitt_potential_2019}
that employs k-medians clustering of the pixels in a scene on multiple scales.\\
\textbf{Sobel Edges}
The coastline detection method presented in~\cite{lee_coastline_1990},
which applies the Sobel filter,
then a spatial dilution process, and then a Roberts edge filter.\\
\textbf{Active Contours}
An active contours approach for coastline detection
based on the Chan-Vese model~\cite{chan_active_1999}.
\subsubsection{Deep Learning}\mbox{}\\
\textbf{HED}
The edge detection model from~\cite{xie_holistically-nested_2015}.\\
\textbf{UNet}
The segmentation model presented in~\cite{ronneberger_u-net_2015},
which is known to work well for coastline detection~\cite{baumhoer_automated_2019}.\\
\textbf{DeepUNet}
A modification of the previous method that was developed for sea-land segmentation
as proposed in~\cite{li_deepunet_2018}.\\
\textbf{RDUNet}
Another modification of UNet developed for sea-land segmentation,
which was proposed in~\cite{shamsolmoali_novel_2019}.\\
\textbf{HRNet + OCR}
One of the current state-of-the-art models for semantic segmentation
in general computer vision~\cite{yuan_object-contextual_2020}.\\
\textbf{Gated-SCNN}
Another recent model for semantic segmentation
in general computer vision~\cite{takikawa_gated-scnn_2019}.
This one is particularly interesting,
as it also combines segmentation with edge detection.

\subsection{Training Details}
The deep learning models were trained on the training dataset of Antarctic coastline scenes
for 15 epochs on a Nvidia V100 card with 32GB of video memory.
The model weights were optimized by an Adam optimizer using the
hyperparameters suggested in~\cite{kingma_adam_2015},
namely a learning rate of $0.001$, $\beta_1 = 0.9$, $\beta_2 = 0.999$ and $\varepsilon = 10^{-8}$.
Due to the large size of the used tiles,
the batch size was set to the low number of 4 samples per batch.

\section{Results and Discussion}
\label{sec:results}
The improved performance from our method 
is quantified using the withheld validation dataset.
To get informative insights on the actual coastline detection performance,
the metrics are calculated only for pixels within \SI{2}{\kilo\metre}
of the true coastline.
This way, a distortion of the metrics from non-coastal areas can be avoided.

The two validation areas
(Antarctic Peninsula and Wilkes Land, see Fig.~\ref{fig:sites})
are evaluated separately.
While the Wilkes Land area can be considered of average difficulty,
the Antarctic Peninsula seems to be a very tough location
for all of the evaluated models.

For the segmentation approaches,
we evaluate the pixelwise accuracy
as well as the mean intersection-over-union metric for the classes of water and land.
For edge detection, we calculate the edge $F_1$ scores at 
optimal image scale (OIS) and optimal dataset scale (ODS)
Finally, we calculate an approximate deviation by averaging the distance
to the ground truth coastline over all predicted coastline pixels (``Deviation'').
Table~\ref{tab:results} shows the numerical results obtained.
The average distance metric can be considered the most important one for this task,
as it estimates the overall error between the actual coastline and the predicted coastline.
Regarding segmentation performance, the mIoU metric can be considered the primary metric.
In order to get a visual impression of some of the models' performance,
Fig.~\ref{fig:qualitative} shows predictions for a selection of validation tiles.
The shown examples are ordered from what we consider easy to hard samples for the models,
and showcase some of the difficulties with the dataset, like sea ice and
confounding backscatter on the higher ice sheet.

\subsection{Model Comparison}
First, it is easy to see that the traditional models
are not really competitive on this dataset.
We ascribe this to the repeatedly stated phenomena
of icebergs and ice sheet regions with difficult backscatter characteristics.
As these models are unsupervised, they simply do not have a way
of learning how to deal with such impediments.

Overall, the heterogeneity of the Antarctic coastline is astounding.
While the coastline is found pretty well by most models in Wilkes Land,
all models have trouble with the scenes from the Antarctic Peninsula.

Among the deep learning based models,
UNet~\cite{ronneberger_u-net_2015} imposes a respectable baseline,
and even outperforms the more recent models
like HRNet+OCR~\cite{yuan_object-contextual_2020} and
Gated-SCNN~\cite{takikawa_gated-scnn_2019} in some of the evaluated metrics.
Even though the latter also has a side output for edge detection,
we find that its edge detection results fall short
in comparison to HED~\cite{xie_holistically-nested_2015} and HED-UNet.
A reason for this might be the lack of a pretrained backbone network
for Sentinel-1 data, which forced us to randomly initialize the backbone
and train it alongside the rest of the network.
Further, this model was optimized for
segmentation of scenes with many different classes and small objects,
which is needed for tasks like autonomous driving.
In our usecase however, there are only two classes
which are nearly equal in area, imposing a very different data distribution.

The ultimate goal of this study is to delineate the coastline
as accurately as possible.
In the corresponding average deviation metric,
the proposed HED-UNet model outshines the alternative approaches,
especially in the Antarctic Peninsula validation area.
This confirms our assumptions that for this specific task,
our considerations lead to increased performance.

\subsection{Network Depth and Deep Supervision}
\begin{table*}
  \center
  \caption{Numerical Results for the Ablations}\label{tab:ablation}%
  \setlength\tabcolsep{3pt}
  \vspace{-2em}
  \scriptsize
  \begin{tabular}{llllllllllllll}
\toprule
             &&& \multicolumn{5}{c}{Wilkes Land} & \multicolumn{5}{c}{Antarctic Peninsula} \\
  Data          &Deep Sup.&Levels&Merging&                       Accuracy &                         mIoU &                            Deviation &                              F$_1$ ODS &                              F$_1$ OIS &                     Accuracy &                         mIoU &                            Deviation &                              F$_1$ ODS &                              F$_1$ OIS \\
\midrule
SAR     &Yes&5&Attention&  90.3$\,\pm\,$1.3 &  82.2$\,\pm\,$2.1 &            239$\,\pm\,$16 &            37.2$\,\pm\,$0.7 &            38.7$\,\pm\,$0.9 &  77.0$\,\pm\,$1.0 &  62.5$\,\pm\,$1.3 &            379$\,\pm\,$45 &            25.4$\,\pm\,$1.2 &            27.0$\,\pm\,$1.0 \\
\midrule
SAR     &No&6&Attention&   88.5$\,\pm\,$1.4 &  79.2$\,\pm\,$2.3 & 954$\,\pm\,$\phantom{0}7 &  \phantom{0}7.1$\,\pm\,$0.5 &  \phantom{0}7.1$\,\pm\,$0.5 &  80.3$\,\pm\,$1.6 &  66.8$\,\pm\,$2.1 &  895$\,\pm\,$\phantom{0}7 &  \phantom{0}7.4$\,\pm\,$0.2 &  \phantom{0}7.5$\,\pm\,$0.2 \\
\midrule                                                                                                                                                                                                                                                                                      
SAR     &Yes&6&None&       89.7$\,\pm\,$1.2 &  81.1$\,\pm\,$1.9 & 284$\,\pm\,$37           &            37.5$\,\pm\,$1.2 &            39.7$\,\pm\,$1.3 &  81.8$\,\pm\,$2.0 &  69.0$\,\pm\,$2.9 &  378$\,\pm\,$35           &            28.0$\,\pm\,$1.8 &            30.2$\,\pm\,$1.8 \\
SAR     &Yes&6&Learned&    89.9$\,\pm\,$2.5 &  81.6$\,\pm\,$3.9 & 236$\,\pm\,$14           &            37.0$\,\pm\,$2.2 &            38.6$\,\pm\,$2.2 &  81.4$\,\pm\,$1.1 &  68.3$\,\pm\,$1.6 &  391$\,\pm\,$12           &            26.6$\,\pm\,$1.4 &            29.1$\,\pm\,$1.4 \\
SAR     &Yes&6&Attention&  92.0$\,\pm\,$0.8 &  84.9$\,\pm\,$1.4 & 222$\,\pm\,$23           &            39.7$\,\pm\,$1.2 &            41.6$\,\pm\,$0.9 &  80.5$\,\pm\,$1.6 &  67.2$\,\pm\,$2.2 &  345$\,\pm\,$24           &            27.1$\,\pm\,$1.9 &            29.4$\,\pm\,$1.8 \\
\midrule                                                                                                                                                                                                                                                                                                         
SAR+DEM &Yes&6&Attention&  92.9$\,\pm\,$1.4 &  86.7$\,\pm\,$2.4 & 226$\,\pm\,$47           &            35.1$\,\pm\,$3.4 &            36.0$\,\pm\,$3.5 &  91.6$\,\pm\,$1.6 &  84.6$\,\pm\,$2.7 &  210$\,\pm\,$\phantom{0}9 &            30.7$\,\pm\,$2.1 &            31.4$\,\pm\,$2.7 \\
\bottomrule
\end{tabular}

\end{table*}
As a means of quantifying the improvements made to the architecture,
we evaluate versions of our model with only some of the improvements applied.
The results of this ablation study are displayed in Table~\ref{tab:ablation}.

For a fair comparison with UNet-based models,
we evaluate the performance when only 5 resolution levels are used instead of 6,
corresponding to 4 down- and upsampling steps instead of 5.
While this setup performs slightly worse than the full HED-UNet,
it still outperforms the baseline methods.

Regarding deep supervision,
we can see that it is of paramount importance for edge detection performance.
Without it, the model is barely able to predict the presence of edges.
What is more, the coastline is often missed completely due to this poor edge detection performance.
On the other hand, deep supervision does not seem
to alter the performance on the semantic segmentation task much.
This is in line with the original models that we took inspiration from.
While the segmentation model UNet~\cite{ronneberger_u-net_2015} does not employ deep supervision,
the edge detection model HED~\cite{xie_holistically-nested_2015} makes heavy use of it.

\begin{figure*}
  \centering
  \includegraphics[width=0.98\textwidth]{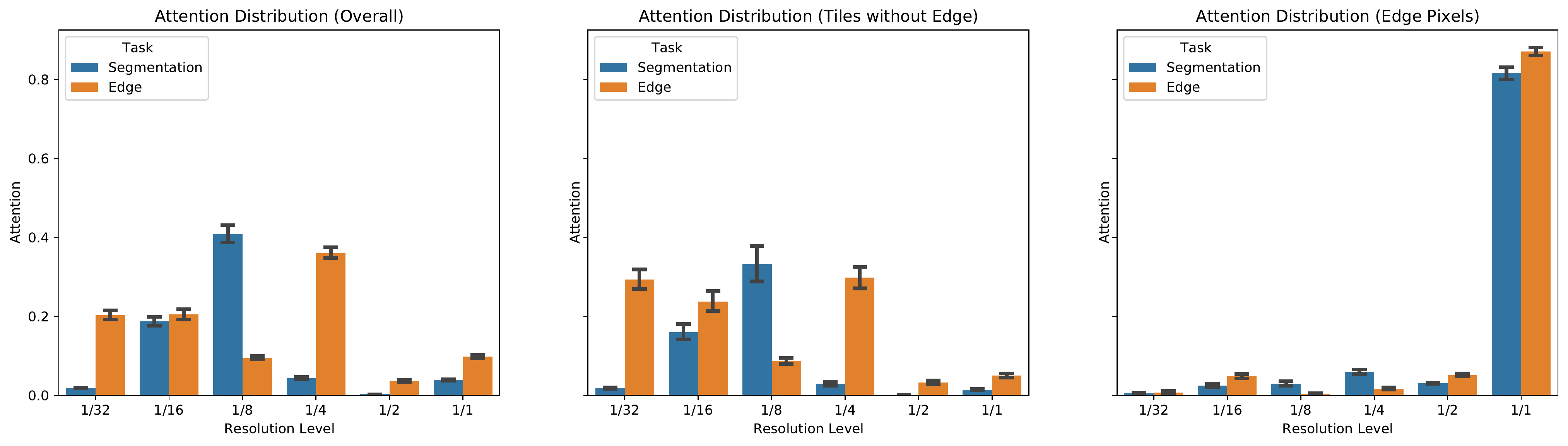}
  \caption{
    Amount of attention spent on the different resolution levels.
    Each plot analyzes a specific class of pixels in the validation dataset --
    from left to right:
    Average over all pixels, 
    average over pixels from edge-less tiles,
    average over all edge pixels.}\label{fig:attn}
\end{figure*}
\subsection{Merging Strategies}
After adding the deep supervision,
we evaluate different merging strategies:
\paragraph{None} First, we evaluate a configuration where just the last layer
  of the decoder is used for the predictions (denoted ``None'').
  This corresponds to the workings of a UNet~\cite{ronneberger_u-net_2015} model
  with two final prediction layers, one for each task.
\paragraph{Learned} Secondly, we evaluate the performance of the learned merging strategy,
  as originally proposed in~\cite{xie_holistically-nested_2015}.
  Here, a prediction is computed for each resolution level in the feature pyramid.
  These predictions are then upsampled to full resolution and concatenated.
  After this, a 1$\times$1 convolutional layer with learned weights computes the final
  prediction from the concatenated prediction stack.
\paragraph{Attention}
  The last strategy is the hierarchical attention merging
  introduced in Sect.~\ref{sec:attn_merging},
  which ]new{does not} rely on fixed weights like the previous strategy,
  but computes the merging weights dynamically for each pixel within each scene.\\[3pt]
  From our results, learned merging does not improve much over no merging
for segmentation, and even performs a bit worse for edge detection.
The average deviation improves quite a bit in Wilkes Land,
but worsens a bit on the Antarctic Peninsula in return.
We ascribe this to the large differences in the validation areas.
As the merging coefficients are fixed for the ``Learned'' approach,
this might hint at the fact that the model learns coefficients that
work well for Wilkes Land, but less so for the Antarctic Peninsula.

This issue is overcome by our newly proposed attention merging strategy,
which can adapt to the different scenes.
It can learn to find good sets of merging coefficients for both Wilkes Land
and the Antarctic Peninsula, even though the optimal values for each one might be different.

Fig.~\ref{fig:attn} shows that the model indeed directs its attention in an
adaptive fashion as we conjectured.
Overall, a mix of all resolution levels is used to compute the final output.
On tiles that are completely covered by one of the two classes,
the attention shifts a bit towards the lower resolution levels,
as they tend to provide more robust predictions.
For pixels on the edge, the model heavily focuses on the
highest available resolution level,
in order to arrive at accurate delineations in these regions.

\begin{figure*}
  \center
  \includegraphics[width=0.8\textwidth]{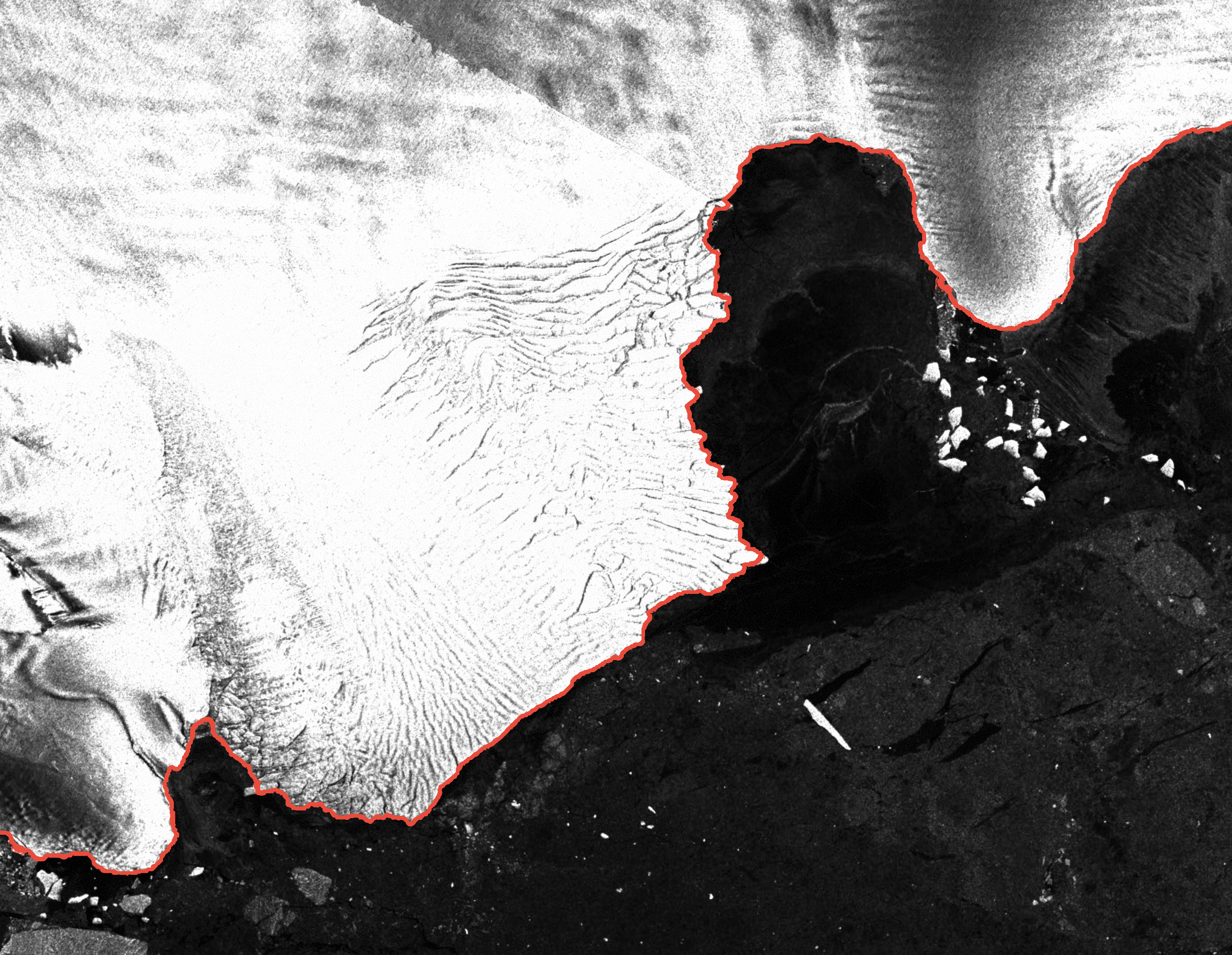}
  \caption{A section of George V Coast with Cape Hudson in the bottom left,
    imagery mosaiced from Sentinel-1 takes in early 2019.
    This scene is both temporally and spatially separated from the training and validation sets used.
    Overlaid in red is the coastline predicted by the HED-UNet model.
  }
  \label{fig:largevis}
\end{figure*}

\begin{figure}
  \centering
  \setlength{\tabcolsep}{1pt}
  \setlength{\tilewidth}{0.15\textwidth}
  \newcommand{\example}[1]{%
    \includegraphics[width=\tilewidth]{figures/autotiles/#1-sar.jpg}&%
    \includegraphics[width=\tilewidth]{figures/autotiles/#1-seg-HED-UNet-S-1.jpg}&%
    \includegraphics[width=\tilewidth]{figures/autotiles/#1-seg-gt.jpg}\\
  }%
  \renewcommand{\arraystretch}{0.55}
  \begin{tabular}{ccccc}
    \textbf{SAR/DEM}&%
    \textbf{HED-UNet}&%
    \textbf{Ground Truth}\\
    \example{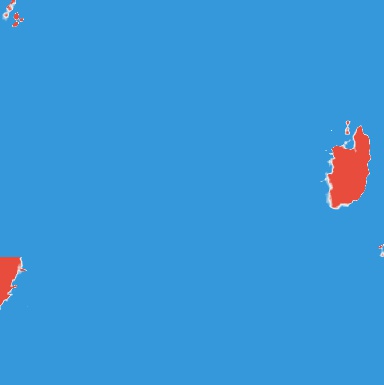}
    \example{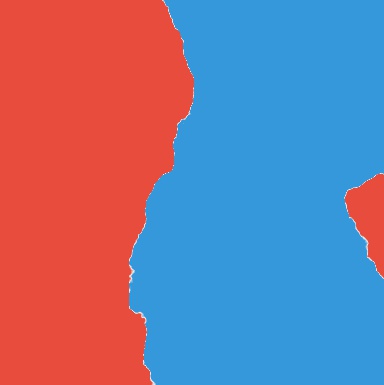}
  \end{tabular}
  \caption{Failure modes of the proposed model.
    Top: Confusion from a very large cluster of sea ice.
    Bottom: Confusion due to missing context at the border of the tile
  }
  \label{fig:failure}
\end{figure}

\subsection{DEM Experiments}
Further, we look into including digital elevation data from the
TanDEM-X elevation model~\cite{rizzoli_generation_2017}.
We conjecture that this secondary data source can help the
model better deject misclassifications from icebergs or
dry-snow facies of the higher ice sheet, which have confounding SAR backscatter.

To discourage the model from directly reproducing the coastline implied by the elevation model,
we decided to downsample the DEM's resolution to \SI{640}{\metre}.
This resolution is coarse enough to not make a segmentation based on the DEM alone
competitive to the non-DEM models, which have an average deviation of less than \SI{300}{\metre}.
Further, it allows for easy feature fusion, as it corresponds to the resolution of the feature
map at $\nicefrac{1}{16}$ of the full resolution.
Therefore, it is simply concatenated to the feature map after
the fourth downsampling step in the encoder.

The results when including the DEM are displayed as the last ablation
in Table~\ref{tab:ablation}.
On the very hard scenes of the Arctic Peninsula,
this additional information helps the model by a large margin,
boosting the average deviation from \SI{345}{\metre} to \SI{210}{\metre}.
However, the story is different for Wilkes Land.
Here, the deviation worsens slightly,
and the edge detection metrics go down considerably.

This is a strong indicator that the model is indeed
overfitting on the DEM to some extent.
For example, in some highly dynamic coastal regions
the model will be confused when the DEM and SAR
imagery are contradictory.

So all in all the inclusion of DEM data can be beneficial,
but needs to be done very carefully to prevent the model from overfitting
to the DEM alone.

\subsection{Limitations}
Even though the newly proposed model outperforms
the baselines on nearly all validation scenes,
there are still cases where the results are not perfect.
Most misclassifications can be attributed to one of two
failure modes, which we will now briefly discuss.
Visual examples for these failure modes can be seen in Fig.~\ref{fig:failure}.

\subsubsection{Sea Ice}
The large receptive field and multitask training help
alleviate the issue of wrongly classified sea ice.
But very large icebergs and areas of ice m\'elange
can still throw off the proposed model.
The first failure example displays such an area where large clusters of sea ice
confuse the model.

\subsubsection{Missing Context}
For areas close to the border of a tile,
the model sometimes does not have enough contextual information to correctly classify them.
This can be observed in the second failure visualization,
where a patch of sea ice directly next to the tile border is wrongly classified as land.

Overall, these failures do not occur often throughout the dataset
and apply not only to the HED-UNet models, but to the other compared models as well.
Especially the first one requires much human interpretation on a large spatial context,
which is difficult for a neural network to achieve without general reasoning capabilities.

\subsection{Effective Receptive Fields}\label{subsec:erf}
Deep CNNs like the ones used in our experiments have very large theoretical receptive fields.
It is conjectured that while long-range connections are theoretically possible in these networks,
networks will often ignore them in favor of short-range connections.

\begin{figure}
  \centering
  \setlength{\tabcolsep}{6pt}
  \begin{tabular}{cc}
    \textbf{UNet~\cite{ronneberger_u-net_2015}}&%
    \textbf{Gated-SCNN~\cite{takikawa_gated-scnn_2019}}\\
    \includegraphics[width=0.45\linewidth]{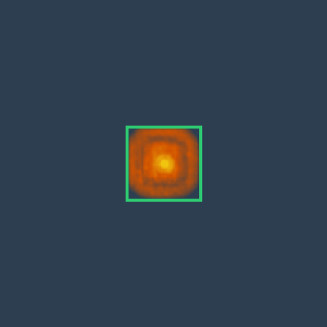}&%
    \includegraphics[width=0.45\linewidth]{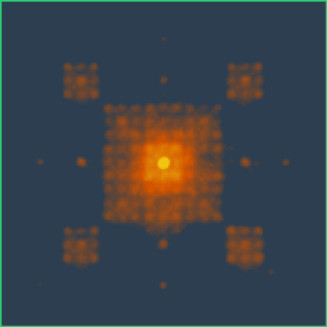}\\
    \textbf{HRNet+OCR~\cite{yuan_object-contextual_2020}}&\textbf{HED-UNet}\\%
    \includegraphics[width=0.45\linewidth]{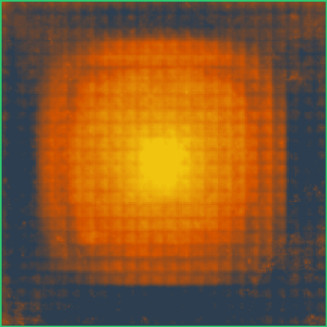}&%
    \includegraphics[width=0.45\linewidth]{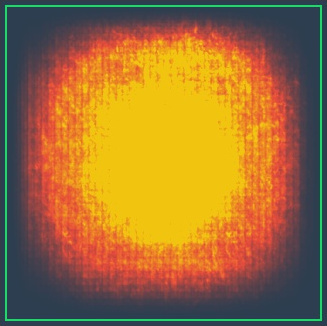}
  \end{tabular}
  \caption{%
    Effective receptive fields of some tested models for the prediction of a central pixel,
    visualized in image space. Theoretical receptive fields outlined in green.
    Note that the theoretical receptive fields of Gated-SCNN and HRNet+OCR
    are larger than the used patch size of 768$\times$768.
  }
\end{figure}

To assess how much of the spatial context is actually used by a CNN,
its so-called \emph{effective receptive field} (ERF)
can be estimated~\cite{luo_understanding_2016}.
This is done by analyzing the expected gradient magnitude of each input pixel
with respect to a central output pixel.
For a CNN $f$ and a sequence of input images $I_k$,
one therefore looks at the values of
\begin{equation}
  E =
  \frac{1}{n}\sum_{k=1}^n
  \left|
  \nabla_{I_k}f(I_k)_{i,j}
  \right|
\end{equation}
for a central output pixel $(i,j)$.
If for an input pixel $(x, y)$, the value $E_{x, y}$ is non-negligible,
then this pixel will influence the output predictions at position $(i,j)$.
The spatial distribution of these relevant pixels is then called the effective receptive field.

As the gradient magnitude gives insight on how much the prediction changes in response
to a change in the input,
the ERF allows for a measurement of the spatial context used by the model.
A model with a larger ERF bases its decisions on a larger spatial context than one with a small ERF.

We conjectured that for the task of Antarctic coastline detection,
a model needs to take a large context window into account.
And indeed, there seems to be a correlation
between a larger ERF and better validation scores for this task.

It can be observed that the UNet model is limited by its theoretical receptive field.
Its ERF is forced into an almost quadratic shape because of this.
The ERF of the Gated-SCNN model is 
  particularly interesting with it's fractal-like shape.
  We conjecture that this is due to the Atrous Spatial Pyramid Pooling block
  used in the network architecture,
  which makes heavy use of dilated convolutions.

Finally, the HRNet+OCR and HED-UNet models employ a very large ERF,
which once more supports our assumption that a large receptive field is
needed for coastline detection in Antarctica.

\section{Conclusion}
In this paper, we introduced a model for simultaneous segmentation and edge detection.
The proposed HED-UNet learns to exploit the synergies between the two tasks,
and thereby manages to surpass both edge detection and semantic segmentation baselines.
By the use of deep supervision, we encourage the model to encode meaningful features
in its deep layers, which allow for more general predictions.
Finally, the proposed attention merging heads allow for
better learning performance and more robust classifications.

Compared to approaching the task with a regular UNet,
the presented network architecture only requires little additional computational cost.
Most of the performance gains stem from the adapted training procedure and a few
additional layers, which do not require many
computational resources compared to the layers already present.

While it is not a general purpose model,
we show that our proposed improvements to the model are indeed beneficial
for the task of coastline detection.
Visual and numerical inspection of the results confirm our assumption that the
combination of the two tasks helps the model better grasp the concept of a coastline.

Our model can be applied to coastline detection tasks
not only in polar regions, but to coastal regions worldwide.
Further, we are convinced that the approach taken by HED-UNet
will greatly benefit other tasks requiring an edge detection approach in combination with
semantic segmentation.
Possible applications include the mapping of building footprints,
roads, and bodies of water like lakes or rivers.
\appendices

\section*{Acknowledgment}
We thank the European Union Copernicus program for providing Sentinel-1.
TanDEM-X elevation data courtesy of the German Aerospace Center (DLR).

\IEEEtriggeratref{60}

\bibliographystyle{IEEEtran}
\bibliography{IEEEfull,references}

\end{document}